%% file: main.tex
\documentclass[10pt,twocolumn,letterpaper]{article}

\usepackage[pagenumbers]{cvpr} 
\usepackage{bbding}
\usepackage{stfloats}
\usepackage{float}
\input{preamble}

\definecolor{cvprblue}{rgb}{0.21,0.49,0.74}
\usepackage[pagebackref,breaklinks,colorlinks,allcolors=cvprblue]{hyperref}

\title{OmniCam: Unified Multimodal Video Generation via Camera Control }

\author{\textbf{Xiaoda Yang}$^1*$ \qquad
\textbf{Jiayang Xu}$^1*$\qquad 
\textbf{Kaixuan Luan}$^1$ \qquad
\textbf{Xinyu Zhan}$^2$ \\
\textbf{Hongshun Qiu}$^3$ \qquad
\textbf{Shijun Shi}$^4$ \qquad
\textbf{Hao Li}$^5$ \qquad
\textbf{Shuai Yang}$^1$ \\
\textbf{Li Zhang}$^5$ \qquad
\textbf{Checheng Yu}$^6$ \qquad
\textbf{Cewu Lu}$^2$ \qquad
\textbf{Lixin Yang}$^2$\footnotemark[2]
\\
$^1$Zhejiang University \qquad
$^2$Shanghai Jiao Tong University \\
$^3$Beijing University of Technology \qquad
$^4$Jiangnan University \\
$^5$University of Science and Technology of China \qquad
$^6$Nanjing university
\\
{\tt\small xiaodayang@zju.edu.cn}
}

\begin{document}

\input{sec/00-teaser}
\footnotetext[1]{*: equal controbution}
\footnotetext[2]{\dag: corresponding author}
\input{sec/0_abstract}
\input{sec/1_intro}
\input{sec/2_relatedwork}

\input{sec/3_dataset}
\input{sec/4_method}

{
    \small
    \bibliographystyle{ieeenat_fullname}
    \bibliography{main}
}

\input{sec/X_suppl}

\end{document}

%% file: preamble.tex
\usepackage{tabularray}
\usepackage{float}
\usepackage{arydshln}
\UseTblrLibrary{booktabs}
\usepackage{booktabs}
\usepackage{multirow}
\usepackage{bm}

\usepackage{graphicx} 
\usepackage{array}

\usepackage{pifont}  

\newcommand{\cmark}{\textcolor[rgb]{0,0.6,0}{\ding{51}}} 
\newcommand{\xmark}{\textcolor{red}{\ding{55}}}   

\usepackage[table]{xcolor}

%% file: sec/00-teaser.tex
\newcommand{\styleFront}{\textbf{\textcolor{frontcolor}{front}}\xspace}
\newcommand{\styleBack}{\textbf{\textcolor{backcolor}{back}}\xspace}
\newcommand{\styleMissingGeometry}{\textbf{\textcolor{sidecolor}{missing geometry}}\xspace}
\newcommand{\styleFaceAndHands}{\textbf{\textcolor{skincolor}{face or hands}}\xspace}

\newcommand{\teaserCaption}{
\textbf {An overview of OmniCam.} Given diverse modalities of content references and trajectory guidance, OmniCam generates high-quality video sequences by camera motion control. Specifically, OmniCam integrates various combinations of content (e.g., image or video) and trajectory (e.g., text instructions or camera motion from video) references. This approach allows OmniCam to accurately synthesize videos consistent with user-specified inputs.}
\twocolumn[{
    \renewcommand\twocolumn[1][]{#1}
    \maketitle
    \centering
    \begin{minipage}{1.00\textwidth}
        \centering
        \includegraphics[trim=000mm 000mm 000mm 000mm, clip=False, width=\linewidth]{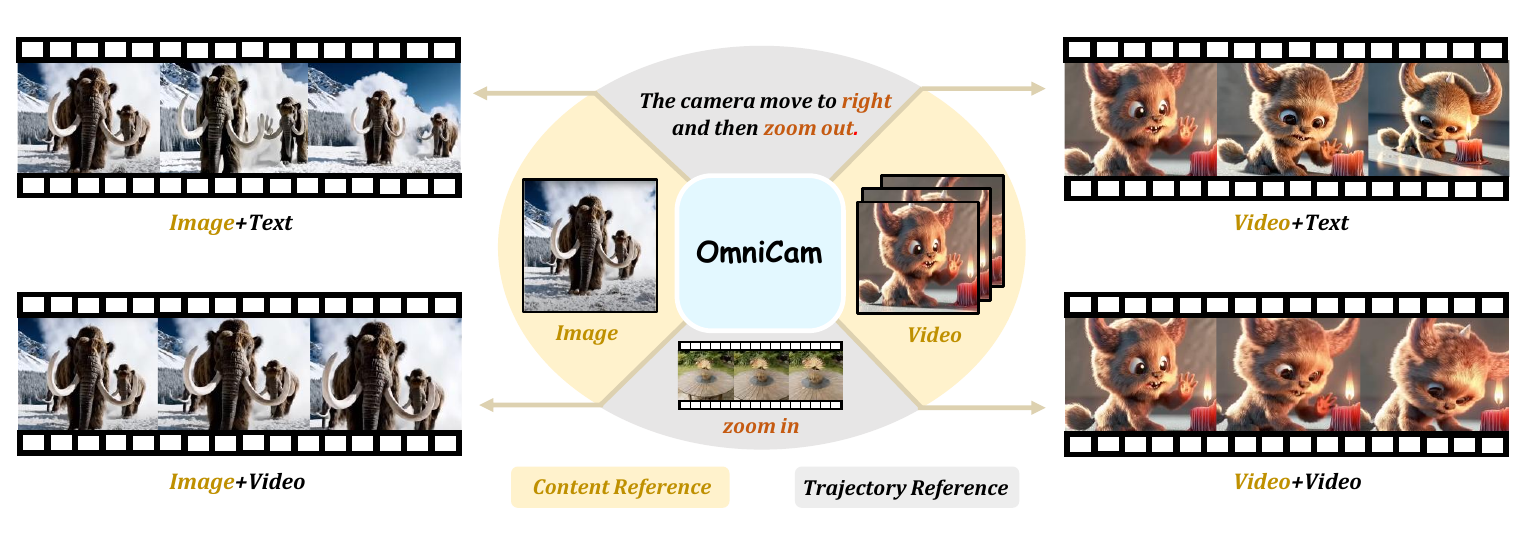}
    \end{minipage}
    \captionsetup{type=figure}
    \captionof{figure}{\teaserCaption}
    \label{fig:teaser}
    \vspace{1.4em}
}]

%% file: sec/0_abstract.tex
\begin{abstract}
Camera control, which achieves diverse visual effects by changing camera position and pose, has attracted widespread attention. However, existing methods face challenges such as complex interaction and limited control capabilities. To address these issues, we present OmniCam, a unified multimodal camera control framework. Leveraging large language models and video diffusion models, OmniCam generates spatio-temporally consistent videos. It supports various combinations of input modalities:  the user can provide text or video with expected trajectory as camera path guidance, and image or video as content reference, enabling precise control over camera motion. To facilitate the training of OmniCam, we introduce the OmniTr dataset, which contains a large collection of high-quality long-sequence trajectories, videos, and corresponding descriptions. Experimental results demonstrate that our model achieves state-of-the-art performance in high-quality camera-controlled video generation across various metrics.
\end{abstract}

%% file: sec/1_intro.tex

\section{Introduction}
\label{sec:intro}
Camera control refers to the task of expressing different visual effects by controlling camera position and pose. The essence of camera movement is a parameterized representation of the four-dimensional space-time continuum, a recognition that is becoming an important breakthrough in spatial intelligence research.

Common technical approaches to camera control include video generation-based methods and reconstruction-based methods. Current development in this field faces several challenges:
I. Existing methods generally cannot perform camera control with a single image input: a) Video generation-based methods cannot distinguish between camera angle changes and subject movement; b) Reconstruction-based methods typically require multi-view information; c) Existing camera control methods often fail to maintain spatial structure, especially during continuous complex operations, resulting in spatial distortion.
II. There is a lack of sufficiently versatile datasets to support these tasks.
III. Current methods have high interaction costs and support only limited modalities. Therefore, we aim to build a 6DoF continuous viewpoint generation method that supports long control sequences and multiple control methods, and to create a sufficiently long and diverse dataset to facilitate this task.

To this end, we propose OmniCam, which first initializes videos following target trajectories through monocular reconstruction, then repairs videos using a video diffusion model with its prior knowledge, and finally adjusts the model through reinforcement learning. It supports frame-level control by setting the start and end frames of operations; supports compound movements in any direction, camera push-pull, allowing movement and rotation to any degree; supports speed control, providing a foundation for quick cuts; supports seamless connection of multiple operations, supporting long sequence operations, allowing continuous execution of multiple instructions; and allows common special effects such as camera rotation. Additionally, our model supports multimodal inputs, consisting of two parts: one providing content information and one providing trajectory information. The content part can be an image or video, and the trajectory part can be provided in two ways: through text descriptions of how the camera moves, or through a video with camera movement effects, from which OmniCam extracts the camera trajectory and applies it to the target video. Beyond this, trajectories can also be input directly, requiring detailed descriptions of camera position at each time step for more refined control, enabling customized camera movement effects.

To train the OmniCam model, we present the OmniTr dataset, the first multimodal camera control dataset, including a large number of multi-stage trajectories and their corresponding videos and text descriptions. Each description includes multiple sub-instructions, with each sub-instruction recording information such as start time, end time, speed, direction, and rotation. Compared to datasets like RealEstate \cite{RealEstate10k}, our dataset includes multiple modalities and provides more diverse trajectory control.

We conducted quantitative and qualitative experiments, demonstrating that OmniCam can achieve flexible and complex trajectory control through multiple modalities. Our contributions are summarized as follows:
\begin{itemize}
\item We propose the OmniCam model, which supports complex and flexible control, allows frame-level control, enables trajectory control through multiple modalities, and processes multi-modal data.

\item We introduce the OmniTr dataset, which collects a large number of high-quality long-sequence trajectories, videos, and descriptions, making it the most modally rich dataset at this scale.

\item We propose a new benchmark for the camera movement field, addressing for the first time issues such as long-sequence trajectory planning in space and video-guided trajectory generation.

\item Experiments show that OmniCam achieves state-of-the-art performance in both quantitative metrics and qualitative effects.
\end{itemize}

%% file: sec/2_relatedwork.tex
\section{Related Work}
\label{sec:formatting}
\subsection{Video Diffusion Models}
\label{2.1}
Recent work on large-scale video diffusion models has achieved high-quality video generation. Video Diffusion Model \cite{VideoDiffusionModels} employs a 3D UNet architecture to jointly learn from images and videos. Imagen Video \cite{ImagenVideo} introduces a cascade structure consisting of seven diffusion models, ingeniously combining key components such as TSR \cite{TSR1,TSR2} and SSR \cite{SSR1,SSR2} for efficient video generation. With the remarkable image quality achieved by text-to-image (T2I) generation models like Stable Diffusion \cite{StableDiffusion}, numerous recent works focus on extending pre-trained T2I models by incorporating temporal modules. Align your latents \cite{AlignYourLatents} proposes a noise map alignment approach, effectively transforming image generation models into video generators. AnimateDiff \cite{AnimateDiff} opts to inject temporal modules into fixed spatial feature layers, creating a framework that enables personalized animation creation without fine-tuning. To enhance temporal consistency, Lumiere \cite{Lumiere} replaces conventional temporal super-resolution modules, directly generating full-frame-rate videos. Recently, researchers have increasingly introduced Transformer \cite{Transformer} architectures into video generation. SORA \cite{Sora} has made significant progress in generating realistic long videos utilizing the DiT \cite{DiT} architecture. Latte \cite{Latte} experiments with four DiT variants for spatiotemporal modeling in latent space, ultimately achieving coherent and realistic generation results.

\subsection{Controllable Video Generation}
\label{2.2}
With the rapid development of generative models across different input modalities, providing diverse guidance for precise control has become a research focus. Recent works such as SparseCtrl \cite{SparseCtrl} and SVD \cite{SVD} utilize images as control signals for video generation. Because of the significance of camera motions in videos, camera control has gained increasing attention. AnimateDiff \cite{AnimateDiff} employs efficient LoRA  \cite{LoRA} fine-tuning to obtain model weights for specific shot types. Direct-a-Video \cite{Direct-a-video} introduces a camera embedder to control camera poses during video generation; however, with three camera parameters, the model supports only basic camera controls like leftward panning. While Runway \cite{RunwayGen2} allows users to freely set camera movements, it is primarily limited to lens-centric operations and suffers from constraints in both movement magnitude and frequency. Regarding trajectory extraction from video modality, although traditional pose estimation algorithms have explored this field, they perform poorly in continuous camera trajectory estimation, particularly in low-frame-rate scenarios. This limitation stems from two technical bottlenecks: First, traditional methods rely on feature point matching algorithms like SIFT \cite{SIFT}, requiring sufficient visual overlap between adjacent frames. In low-frame-rate scenarios, large camera movements often lead to feature-matching failures. Second, increased inter-frame intervals significantly affect the prediction accuracy of motion models, increasing uncertainty in motion estimation. Our approach not only supports text-based guidance but also effectively extracts camera trajectories from target videos in low-frame-rate environments, providing a more flexible and robust solution for controllable video generation.

\begin{figure*}[!htbp]
\centering
\includegraphics[width=\textwidth]{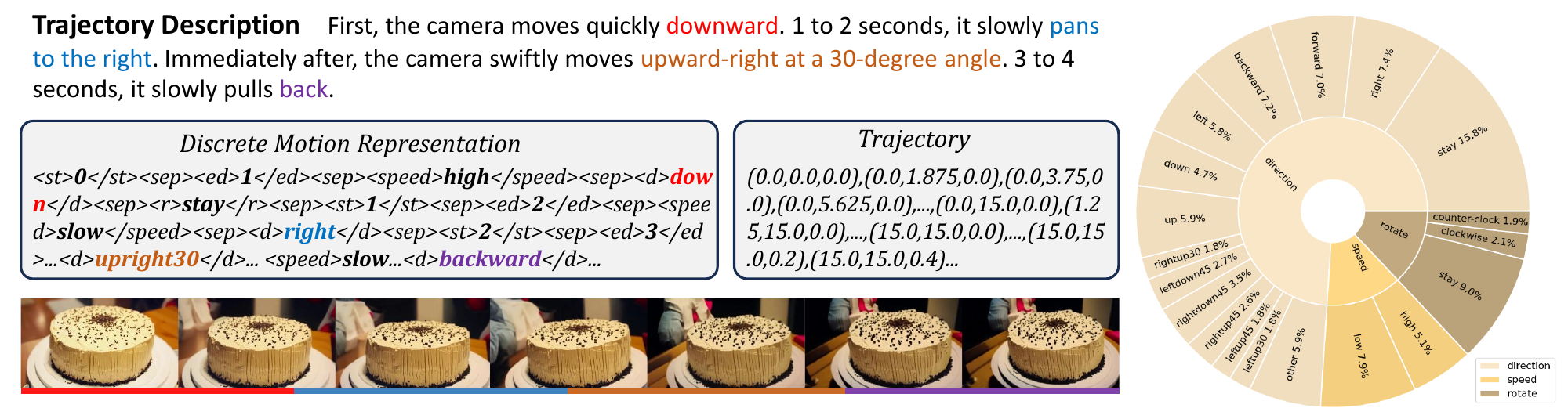} 
\caption{\textbf{OmniTr dataset consists of four key components:} trajectory description, discrete motion representation, trajectory, and corresponding video sequence. Notably, we visualized the discrete motion representations, with the pie chart on the right clearly showing the distribution proportions of various motion attributes. As can be seen, our dataset comprehensively covers all motion attributes.}
\label{fig:dataset}
\end{figure*}

%% file: sec/3_dataset.tex
\section{OmniTr Dataset}
\label{sec:formatting}
Existing datasets lack long-sequence camera control and cannot support complex, flexible camera movements and multimodal input. To address these issues, we introduce the OmniTr dataset, a large-scale resource library specifically designed for comprehensive camera control.

As shown in Fig. \ref{fig:dataset}, OmniTr uses trajectory groups as its basic unit. Each group contains four components: trajectory description, discrete motion representation, trajectory in polar coordinates, and high-quality videos. We carefully constructed 1000 unique trajectory groups, generating a comprehensive camera control dataset containing 1000 trajectories, 10,000 descriptions, 30,000 videos, and their corresponding discrete motion representations. Among them, each trajectory corresponds to 10 descriptions and 30 videos of common categories.

OmniTr builds its video content based on the CO3D dataset and uses large language models to generate diverse textual descriptions. The dataset provides control information accurate to the frame level, with discrete motion representations that can be directly converted into complete six-degrees-of-freedom (6DoF) sequences, enabling precise camera trajectory control.

As shown in Tab. \ref{tab:datasets}, we compare the OmniTr dataset with existing datasets. Our dataset is not only large-scale but also comprehensively covers all camera control methods.
Each text description consists of 1 to 5 camera operation descriptions, with each operation associated with a specific discrete motion representation. The text precisely expresses the time range, speed, direction, and angle of operations. The discrete motion representation contains several key fields: \textbf{starttime} and \textbf{endtime} describes the operation time period; \textbf{speed} describes the camera movement speed (low indicates slow, high indicates fast); \textbf{direction} describes the movement direction (including basic directions and combination directions at any angle); \textbf{rotate} describes the rotation method (clockwise, counterclockwise, or stationary). The pie chart in Fig. \ref{fig:dataset} demonstrates the dataset's comprehensive coverage of various operation methods.

To enhance the dataset's utility, we improve the text quality in multiple dimensions.
\textbf{Time robustness processing}: Some operations explicitly specify a time range (such as “0 to 1 second”), while for others without explicitly specified times, default values are used: camera movement defaults to 1 second, and camera rotation defaults to 0.5 seconds. Operation times in some texts are non-continuous; for example, the first operation occurs at 0-1 seconds, while the second operation jumps to 3-4 seconds, with no additional operations during the intermediate time;
\textbf{Angle flexibility processing}: The input text allows for any angle in combination directions;
\textbf{Language styles diversifying}: We include formal statements, concise expressions, exaggerated descriptions, and other forms to ensure rich semantics and adaptability to different scenario requirements.

\begin{table}[t]
\centering

\renewcommand{\arraystretch}{1.3} 

\Huge
\resizebox{\columnwidth}{!}{
\begin{tabular}{@{}cccccccccc@{}} 
\toprule
\rowcolor[rgb]{ .961,  .961,  .961}
\textbf{Datasets}      & \textbf{Modality} & \textbf{Any-direction} & \textbf{Zoom} & \textbf{Rotate} & \textbf{Speed} & \textbf{Complex} \\ 
\midrule
Tanks\&Temples \cite{Knapitsch2017}           & Video                & \cmark & \cmark & \xmark & \xmark & \cmark \\ 
RealEstate10k \cite{zhou2018stereo}             & Video                 & \cmark & \cmark & \cmark & \xmark & \cmark \\ 
CO3D \cite{co3d}                   & Image                 & \xmark & \xmark & \xmark & \xmark & \xmark \\ 
Webvid10m \cite{Bain21}            & Video                 & \xmark & \xmark & \xmark & \xmark & \cmark \\ 
OmniTr                   & T+V                & \cmark & \cmark & \cmark & \cmark & \cmark \\ 
\bottomrule
\end{tabular}
}
\caption{\textbf{Comparison of other datasets with OmniTr.} None of the other datasets include textual descriptions of the trajectories. T in the table stands for Text.}
\label{tab:datasets}
\end{table} 

%% file: sec/4_method.tex
\begin{figure*}[t]
\centering
\includegraphics[width=\textwidth]{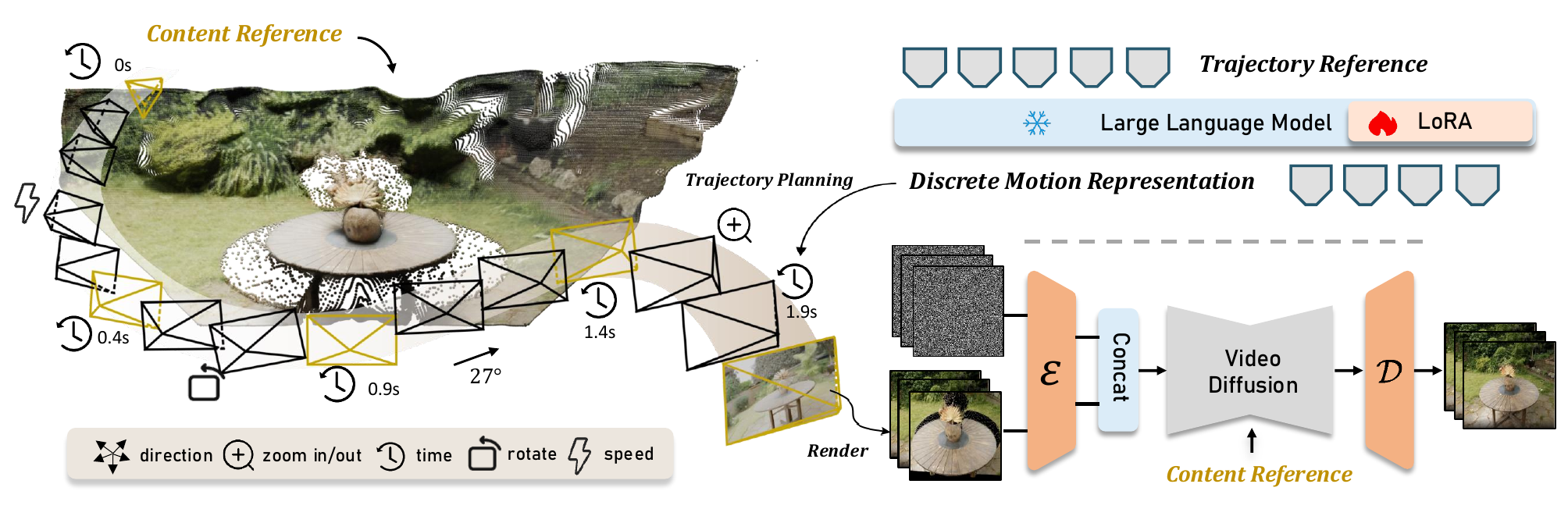} 
\caption{\textbf{An overview of OmniCam model pipeline.} After receiving the trajectory reference, OmniCam first converts it into discrete motion representations through LLM. Subsequently, OmniCam uses a trajectory planning algorithm to calculate the camera pose for each frame based on these motions. Combined with the content reference, OmniCam renders the initial view for each frame. Finally, it utilizes a diffusion model to complete unknown regions in the new viewpoints, and stitch all frames together to generate a coherent video.}
\label{fig:3}
\end{figure*}

\section{Method}
\subsection{Trajectory Generation}
\label{4.1}
Our camera trajectory generation system aims to produce appropriate trajectories based on input descriptions.
Previous works are typically confined to single-dimensional camera movements or limited to simple camera operations at key frames, while struggling with both precise control and long-range trajectory planning.
Our method achieves frame-level precision through fine-grained control, supporting compound movements in arbitrary directions and camera zoom operations. Furthermore, our approach enables seamless integration of multiple operations, resulting in more natural and fluid camera trajectories. The system accommodates multi-modal inputs, accepting both text and video sequences as trajectory references.

\subsubsection{Description to Discrete Motion Representation}
We utilize discrete motion representation as an intermediate representation for generating continuous trajectories, facilitating long-range fine-grained control. For textual inputs, a large language model is used to transform description into discrete motion representation, comprising sequences of $<$starttime, endtime, speed, direction, rotate$>$. 

We fine-tune the large language model using LoRA (Low-Rank Adaptation). The input sequence can be represented as $\mathit{Y}=\{y_1, y_2, ..., y_k\}$. The discrete motion representation comprises multiple primitive statements, each containing a quintuple of control signals and separated by $_<sep_>$ tokens. The resulting discrete motion representation sequence is formulated as $\hat{Y}=\{\hat{y_1}, \hat{y_2}, ..., \hat{y_t}\} $, where $t \in R^N$ . Here, $N$ denotes the number of control signals derived from the description and satisfies the condition $5|N$. The loss function for discrete motion representation prediction is defined as:

\begin{equation}
\mathcal{L}_\mathrm{trajectory} = -\sum_{t=1}^{T}\log p(y_t|\hat{y}_{<t})
\end{equation}

where $T$ denotes the total length of the trajectory sequence. 
$p(y_t|\hat{y}_{<t})$ represents the probability of correct model prediction and $\hat{y}_{<t}$ indicates all predicted results prior to time step $t$.

Similarly, video-guided camera control also utilizes discrete motion representation as a bridge. While trajectory extraction from videos has been explored in camera pose estimation research \cite{extractTr1,extractTr2,extractTr3,extractTr4,extractTr5,extractTr6}, traditional camera pose estimation methods primarily focus on reconstruction assistance rather than optimizing continuous camera trajectory prediction, leading to abrupt and inconsistent results when processing video, particularly in low-frame-rate scenarios. To address this issue, we incorporate a smoothing module following the feature extractor, which maps trajectories to the downstream coordinate system while ensuring continuity and smoothness. Compared to traditional camera pose estimation approaches, our method demonstrates robust performance even in low-frame-rate conditions.

\subsubsection{Discrete Motion Representation to Trajectory}
After obtaining the discrete motion representation,
 we employ the \textbf{trajectory planning algorithm} to calculate the spatial position of each point in the trajectory. The spatial position is parameterized by a triplet $(\phi,\theta,r)$, which includes the azimuth angle $\phi$, polar angle $\theta$, and radius $r$. The rotation is independently controlled by the rotate parameter in the discrete motion representation. These positions are then converted into a sequence of camera extrinsic parameters. This process is lossless and reversible. This algorithm models the camera movement around the object's center as a spherical motion by default. Initially, we compute the start and end frames affected by the control signals. Subsequently, based on specific control information, we calculate an incremental change $\delta =F(v, d)$, where $v$ and $d$ represent the control information for velocity and direction, respectively. The function accumulates corresponding angular or distance changes $\Delta=\Sigma \delta_t$ for each frame to generate the complete trajectory sequence.

Specifically, given a motion instruction, we first calculate the number of frames based on the frame rate and duration. Let $f$ denote the frame rate and $t$ represent the total duration; the total number of frames is expressed as $f \times t $. Through this approach, utilizing the concept of division points in a given ratio \cite{divisionPoint}, we unify temporal and frame-based control. To simulate velocity variations, we initialize a unit incremental change $\delta^{v}$  and apply scaling factors, supporting control granularity at both high and low levels. For translational operations, the pose of each frame is computed by adding the increment to the pose of the previous frame, thereby forming a complete pose sequence for the corresponding operation. Rotational operations are implemented according to the details in the supplementary.
\subsection{Trajectory-guided Video Synthesis}
\label{4.2}

There are various methods for generating videos from trajectories. Traditional approaches typically utilize spatial transformer block to capture content information and temporal transformer block to handle temporal dependencies, injecting trajectory information as conditions into the model. However, the performance of such model architectures has been limited. Our method opts to leverage 3D reconstruction to obtain fundamental content information and temporal dependencies. While existing reconstruction methods often rely on 3D Gaussian Splatting \cite{3DGS} this technique involves a complex processing pipeline: it requires locating the Gaussian ellipsoid's center from point clouds, calculating the covariance matrix to construct the ellipsoid, adding opacity information, and finally rendering the video according to the target trajectory. Given the complexity of this process, we have chosen to directly utilize point clouds for monocular reconstruction and introduce a diffusion model to resolve the unknown regions in the rendering.

\subsubsection{Reconstruction and Render}
Point clouds, camera intrinsic parameters, and camera extrinsic parameters are required simultaneously during the rendering process. Point clouds and camera intrinsics are acquired through DUSt3R \cite{Dust3r}, with camera extrinsics obtained from LLM outputs. Specifically, DUSt3R takes two images as input, computes a disparity map through Dense Stereo Matching \cite{Stereo_Matching}, and generates point cloud data. In addition, confidence maps are used to assist in subsequent intrinsic parameter estimation. The Weiszfeld algorithm \cite{Weiszfeld} is then used to optimize the camera focal length \( f_0 \), under the assumption that the principal point is located at the center of the image and that the pixels are square. This completes the estimation of the camera's intrinsic parameters. When there is only a single input image, it can be copied to construct a paired input for compatibility with DUSt3R.


For image \( I_\text{ref} \) serving as content references, we first use the dense stereo model \cite{Dust3r} to extract its point cloud data, camera intrinsic parameters, and camera pose \( C_\text{ref} \). Next, navigate the camera using the camera pose sequence \( C = \{C_0, \dots, C_{L-1}\} \) (including \( C_\text{ref} \)), render the point cloud, and generate a series of rendering results \( P = \{P_0, \dots, P_{L-1}\} \). Our goal is to learn the conditional distribution \( x \sim p(x | I_\text{ref}, P) \) and generate high-quality perspective conversion videos \( x = \{x_0, \dots, x_{L-1}\} \) by rendering point cloud \( P \) and reference image \( I_\text{ref} \).

For videos serving as content references, we perform the same processing workflow frame by frame. For instance, when inferring the i-th frame of the input video, we select the i-th frame from the inferred video, and so forth, assembling the results frame by frame. To enhance generation efficiency, we conduct inference at regular intervals to obtain key frames, which are then assembled, and employ frame interpolation techniques to increase the video's fluidity.
\subsubsection{Resolve the Unknown Region}

As shown in Fig. \ref{fig:3}, point cloud rendering results typically contain unknown regions. Similar to how humans can imagine the back of an object based on its front view, diffusion models also possess this imaginative capability based on prior knowledge, which is why we apply them to complete these unknown regions.

To improve computational efficiency, we employ a latent diffusion model (LDMs) \cite{LDM} architecture, using a pre-trained VAE to map images into a low-dimensional latent space for processing. Inspired by previous work \cite{DynamiCrafter,ViewCrafter}, we construct a high-quality paired dataset containing point cloud rendering sequences \( P = \{P^0, \dots, P^{L-1}\} \) and corresponding real reference images \( I = \{I^0, \dots, I^{L-1}\} \).

During training, we freeze the VAE \cite{VAE,VAE0} encoder-decoder parameters and focus on optimizing the latent space. Specifically, we encode the training data $I$ and $P$ into latent variables \( z = \{z^0, \dots, z^{L-1}\} \) and condition signals \( \hat{z} = \{\hat{z}^0, \dots, \hat{z}^{L-1}\} \), concatenating them with sampled noise \( \epsilon \sim \mathcal{N}(0, I) \) along the channel dimension before input to the U-Net. The model is optimized using the following diffusion loss function:
\begin{equation}
\min_\theta = \mathbb{E}_{t \sim \mathcal{U}(0, 1), \epsilon \sim \mathcal{N}(0, I)} \left[\|\epsilon_\theta(z_t, t, \hat{z}, I_\text{ref}) - \epsilon\|^2 \right]
\end{equation}
where \( z_t = \alpha_t z_0 + \sigma_t \epsilon \).

Additionally, we inject the CLIP \cite{CLIP} features of the reference image as conditions into the UNet \cite{Unet} to prevent domain shift.

During inference, we first render the point cloud sequence, and then encode the obtained image sequence into latent variables and concatenate them with noise samples.
Subsequently, we use the trained U-Net to denoise the latent variables iteratively. Finally, we transform the results into high-fidelity view-transformed outputs through the VAE decoder, achieving a transition from incomplete rendering to complete visual effects.

\subsection{End-to-End Optimization}
To enhance system performance, after completing the pre-training of the two modules described in Sec. \ref{4.1} and Sec. \ref{4.2}, we further explore coupling optimization between them. Due to the gradient truncation caused by token outputs from the upstream module, we adopt a reinforcement-learning-based approach to achieve end-to-end optimization, improving the coupling effectiveness of the model.
Specifically, we treated the downstream network as a reward model, using downstream feedback to guide the optimization of upstream modules. This feedback mechanism established effective interaction between upstream and downstream modules, significantly enhancing the collaborative performance of the overall model, allowing downstream evaluation results to directly influence and improve upstream generation processes.

Our implementation resembles the RLHF (Reinforcement Learning from Human Feedback) \cite{RLHF} framework: first, we freeze the downstream model and use it as a reward function; then, we employ the trajectory extractor from Sec. \ref{4.1} to obtain camera pose sequences of generated videos, using the average of various dimensional scores as rewards to update the upstream model. Simultaneously, to prevent excessive deviation in model capabilities, we established a reference model that was initialized from the upstream module and kept frozen throughout training, which ensures the model retains its original performance while acquiring new capabilities.

\section{Experiment}
\subsection{Implement Details}
OmniCam employs a three-stage training strategy. In the first stage, a large-scale model along with its associated components is trained. Llama3.1 \cite{llama1,llama2} serves as the backbone, which is subsequently fine-tuned with Lora. In the second stage, the video diffusion model is trained with a learning rate of 5e-5 and a batch size of 16, undergoing 50,000 iterations. The training is conducted on RealEstate \cite{RealEstate10k}, DL3DV \cite{Dl3dv}, and a selected subset of the OmniTr datasets, with each video comprising 25 frames. During the inference process, we employ the DDIM \cite{DDIM} sampler and classifier-free guidance. In the third stage, we fine-tune the trajectory large model using reinforcement learning, freeze the downstream video generation model, and employ it as a reward model, leveraging PPO \cite{PPO} to fine-tune the large model. We train the model and variants on 8 NVIDIA A100 GPUs. More specific implementation details can be found in the supplementary.

\begin{table*}[ht]
\renewcommand{\arraystretch}{1.2}
\resizebox{\textwidth}{!}{
\begin{tabular}{llccccccclclc}
\hline
\multirow{2}{*}{Modality}                   & \multirow{2}{*}{Backbone} & \multicolumn{7}{c}{Trajectory}                                                                        &  & Quality             &  & Consist             \\ \cline{3-9} \cline{11-11} \cline{13-13} 
                                            &                           & $Avg$  & $M_{d-course}$ & $M_{d-fine}$ & $M_{speed}$ & $M_{rotate}$ & $M_{starttime}$ & $M_{endtime}$ &  & NIQE $\downarrow$ &  & CLIPSR $\uparrow$ \\ \cline{1-13} 
\hline
\rowcolor[HTML]{E8F8F5}
Image(C)+Text(T)                            & Llama                     & \textbf{80.171} & \textbf{85.267}         & \textbf{76.488 }      & \textbf{79.818}      & \textbf{79.818 }      & \textbf{79.818  }        & \textbf{79.818  }      &  & 2.831             &  & 0.925               \\
\rowcolor[HTML]{E8F8F5}
Image(C)+Text(T)                            & Qwen2-VL                  & 72.976 & 83.251         & 64.519       & 70.763      & 71.655       & 74.827          & 72.844        &  & \textbf{2.830}               &  & \textbf{0.926 }              \\ 
\rowcolor[HTML]{F4ECF7}
Image(C)+Video(T) & Llama + SLAM              & \textbf{45.667} & \textbf{68.667 }        & \textbf{6.833 }       & \textbf{28.000 }     &\textbf{ 52.583  }     & \textbf{62.167  }        & \textbf{55.750 }       &  & \textbf{2.831 }              &  & 0.923               \\
\rowcolor[HTML]{F4ECF7}
Image(C)+Video(T) & SLAM              & 36.350 & 59.562         & 6.225        & 21.151      & 41.276       & 44.123          & 43.767        &  & 2.831               &  & 0.923               \\
\rowcolor[HTML]{F4ECF7}
Image(C)+Video(T)                           & Llama + SIFT              & 23.445 & 33.212         & 1.128        & 14.352      & 28.438       & 32.415          & 31.127        &  & 2.831               &  & 0.925               \\
\rowcolor[HTML]{F4ECF7}
Image(C)+Video(T)                           & Qwen2-VL                  & 33.931 & 56.232         & 5.245        & 19.361      & 37.279       & 43.121          & 42.353        &  & 2.831               &  & \textbf{0.926 }              \\ 
\rowcolor[HTML]{FDF8E7}
Video(C)+Text(T)  & Llama                     & 80.171 & 85.267         & 76.488       & 79.818      & 79.818       & 79.818          & 79.818        &  & 3.796               &  & 0.940               \\ 
\rowcolor[HTML]{EBF5FB}
 Image(C)+Text(T)  & Llama (w/o RL)            & 78.341 & 83.153         & 74.362       & 78.133      & 78.133       & 78.133          & 78.133        &  & 2.831               &  & 0.925               \\ \hline
\end{tabular}
}
\caption{The performance of different modalities (including content reference and trajectory reference) under different backbones across multiple indicators. These metrics are used to determine the effectiveness of extracting tracks from text or video. The best result is \textbf{bolded}.}
\label{tab:ablation}
\end{table*}

\begin{table}[h]
\renewcommand{\arraystretch}{1.17}
\resizebox{1\columnwidth}{!}{
\begin{tabular}{ccccccc}
\hline
\rowcolor[rgb]{ .961,  .961,  .961} Method       & RotErr↓  & TransErr↓  & LPIPS↓ & PSNR↑ & FID↓  \\ \hline
CameraCtrl \cite{CameraCtrl}   & 6.423                                   & 5.792                                   & 0.291  & \underline{18.37} & 69.4 \\
LucidDreamer \cite{liang2024luciddreamer} & 7.990                                   & 10.85                                   & 0.400  & 14.13 & 71.43 \\
CamI2V \cite{CamI2V}    & \underline{5.983}                                   & \underline{5.221}                                   & \underline{0.267}  & 18.27  & \underline{58.30} \\
ZeroNVS \cite{zeroNVS}     & 8.562                                   & 10.31                                   & 0.431  & 14.24  & 105.8 \\
MotionCtrl \cite{wang2024motionctrl}   & 8.084                                   & 9.295                                  & 0.386  & 16.29  & 70.02 \\
OmniCam (Ours)      & \textbf{1.066}                                   & \textbf{2.731}                                   & \textbf{0.167}  & \textbf{22.14}  & \textbf{24.26} \\ \hline
\end{tabular}
}
\caption{The comparison with other models. More comparisons with other methods can be found at supplementary and demo page.}
\label{tab:compare_with_other_method}
\end{table}


\subsection{Metrics}
To evaluate the accuracy of the generated trajectories, we propose five metrics to supervise the discrete motion representation. $M_{starttime}$ and $M_{endtime}$ are used to evaluate the model's accuracy in determining the start and end times. $M_{speed}$ is employed to assess the model's understanding of speed. $M_{rotate}$ is utilized to evaluate whether the model correctly comprehends the direction of rotation. $M_{direction}$ is used to judge the model's ability to accurately understand the direction of camera movement. All these metrics are essentially accuracy measures, calculated as the average of all subtasks. Among these metrics, $M_{starttime}$ and $M_{endtime}$ are evaluated independently, while $M_{rotate}$, $M_{speed}$, and $M_{direction}$ are assessed based on the condition that both the start time and end time are correctly determined. Additionally, due to the complexity of directional information, we divide $M_{direction}$ into $M_{d-course}$ and $M_{d-fine}$. $M_{d-course}$ allows for discrepancies in degrees but requires the direction to be correct, while $M_{d-fine}$ demands both the direction and degrees to be accurate. Since the discrete motion representation uniquely determines the trajectory, these metrics can evaluate the accuracy of the model-generated trajectories. We are the first to propose evaluation criteria for extracting complex trajectories from text or video, thereby laying a foundation for future research.

Additionally, we evaluate the generated videos using several classic metrics, such as LPIPS \cite{LPIPS}, PSNR \cite{PSNR}, SSIM \cite{SSIM1,SSIM2}, FID \cite{FID}, NIQE \cite{NIQE}, and CLIPSR \cite{CLIPScore}. These metrics assess the quality and smoothness, and more details can be found in the supplementary materials.

We then calculate the rotation distance (RotErr) in comparison to the ground truth rotation matrices of each generated novel view sequence, expressed as:
\begin{equation}
    \mathrm{RotErr} = \sum_{i=1}^{n} \arccos\frac{tr(\bm{r}_{\text{gen}}^i \cdot \bm{r}_{\text{gt}}^{iT}) - 1}{2},
\end{equation}

where $\bm{r}_{gt}^{i}$ and $\bm{r}_{gen}^{i}$ denote the ground truth rotation matrix and generated rotation matrix. We also compute the translation distance (TransErr), expressed as:
\begin{equation}
    \mathrm{TransErr} = \sum_{i=1}^{n} \|\bm{t}_{\text{gt}}^i - \bm{t}_{\text{gen}}^i\|_2,
\end{equation}

where $\bm{t}_{gt}^{i}$ and $\bm{t}_{gen}^{i}$ denote the ground truth translation matrix and generated translation matrix. 
Since COLMAP \cite{colmap} is sensitive to inconsistent features \cite{ViewCrafter}, we use DUSt3R \cite{Dust3r} for more robust pose estimation.

\begin{figure*}[!htbp]
\centering
\includegraphics[width=0.98\textwidth]{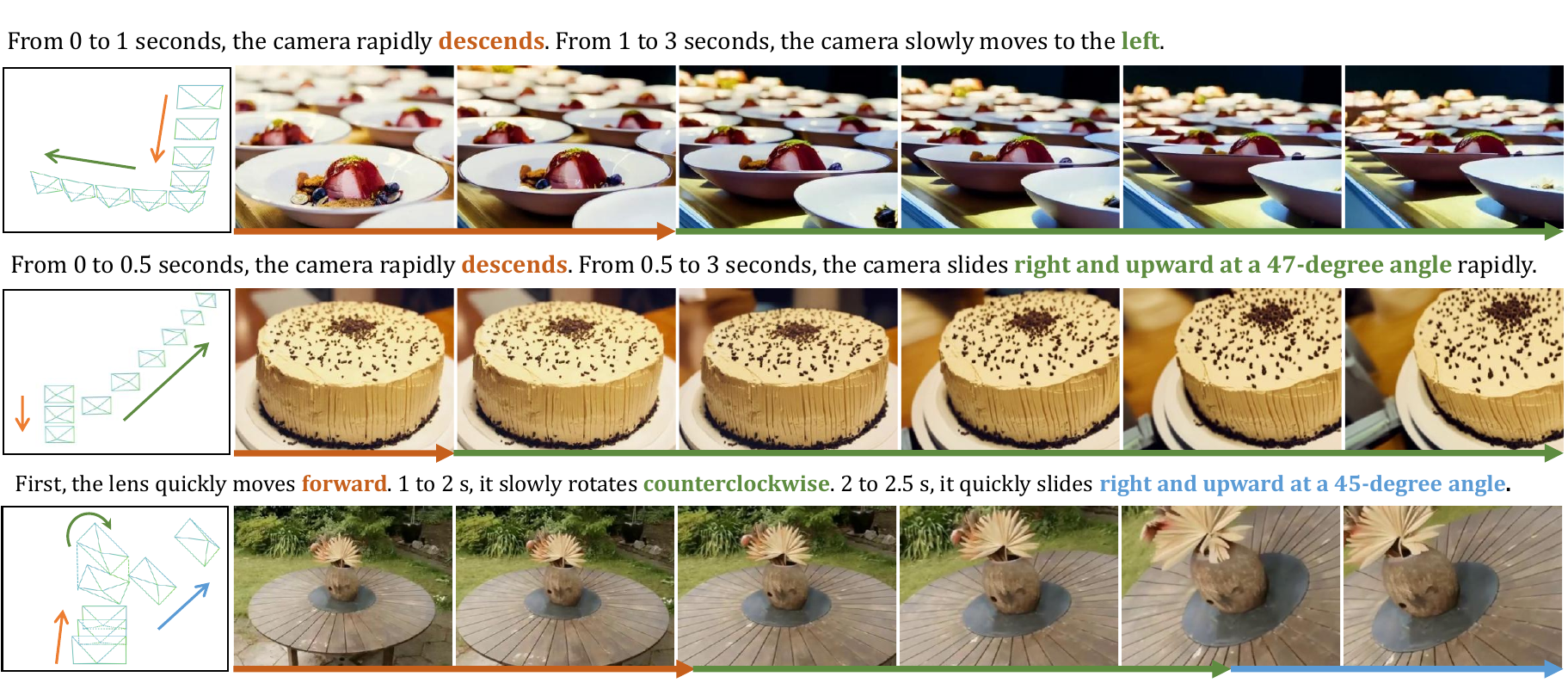} 
\vspace{-0.3em}
\caption{\textbf{Text description  for camera control}. Each set of results demonstrates the generation effects of different types of camera motion combinations, including directional movements at specified angles, rotations, and other complex movements.}
\label{fig:demo1}
\end{figure*}

\begin{figure*}[!htbp]
\centering
\includegraphics[width=0.98\textwidth]{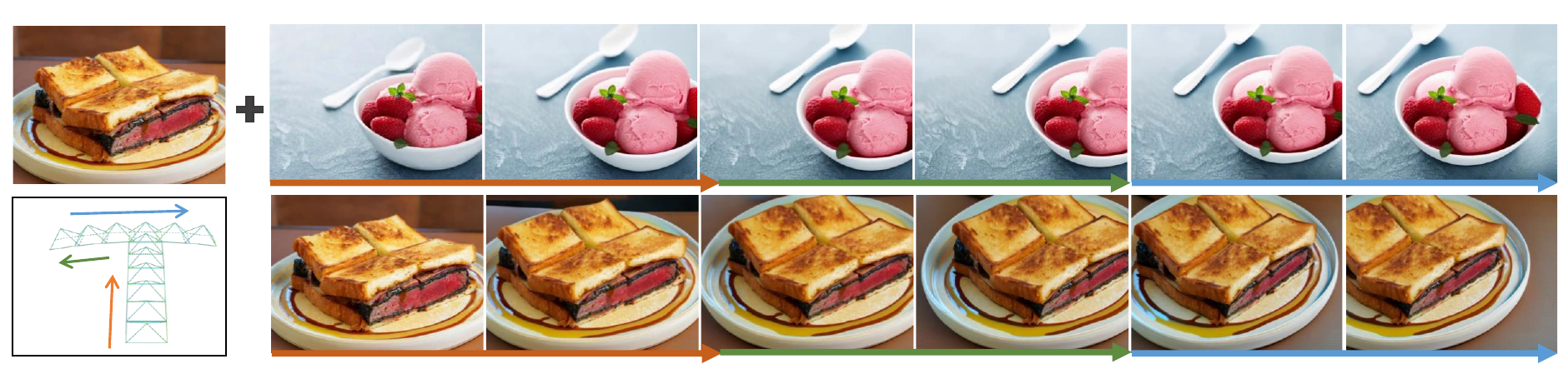} 
\vspace{-0.3em}
\caption{\textbf{Video trajectory for camera control}. OmniCam transfers the trajectory extracted from the input video to the output video. The first line represents the input and the second line represents the output.}
\label{fig:demo2}
\end{figure*}

\subsection{Main Results and Ablation Study}
As shown in Tab. \ref{tab:ablation}, for the task of obtaining trajectories through text descriptions, we compare two backbone models. The experimental results demonstrate that the LLM performs better. This is because the LLM is more focused on text comprehension, demonstrating superior understanding capabilities compared to the VLM (Vision Language Model). For the task of extracting trajectories from videos, we compared three different approaches. The experimental results show that the Llama + SLAM \cite{SLAM} approach performs the best. This is because camera pose estimation requires strong three-dimensional spatial perception, which current VLM models still lack. Therefore, we adopt a mature pose estimation algorithm as the foundation, and used an LLM combined with an MLP as a mapper to correct and adjust the generated trajectories. In comparison, the SIFT \cite{SIFT} method offers faster generation speed, but the results are not satisfactory. Building upon the optimal training, we utilized offline policy reinforcement learning. 
\begin{figure}[H]
\includegraphics[width=\columnwidth]{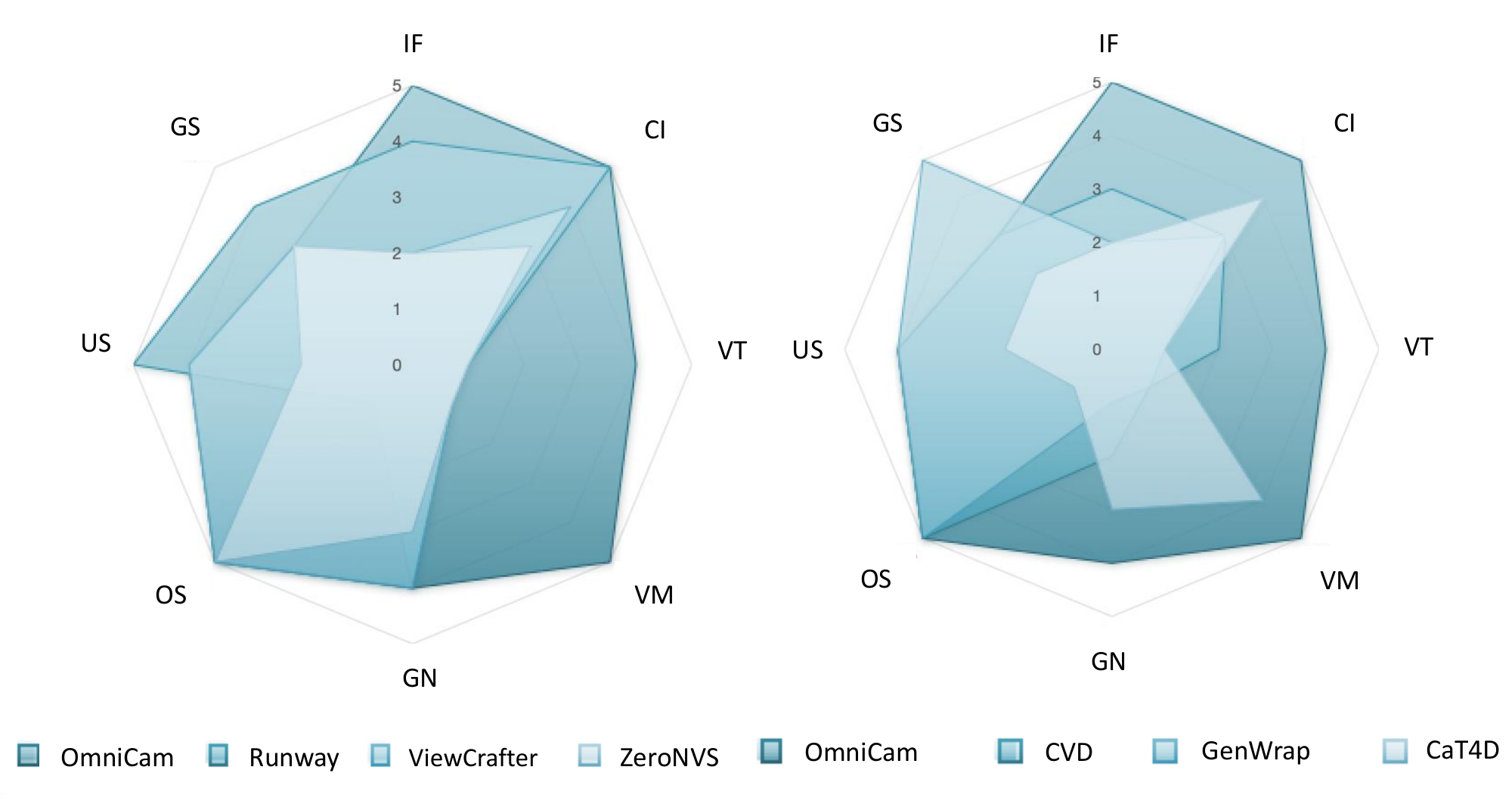} 
\caption{Radar chart comparing OmniCam with domain-specific models across eight dimensions: IF stands for Integration Flexibility, OS refers to Open Source, GS means Generation Speed, VM represents Video Manipulation, GN is Generalization, US indicates Usability, CI denotes Support Complex Instructions, and VT signifies Video Trajectory-Based. 
}
\label{fig:radar}
\end{figure}
\vspace{-5pt}
The experimental results indicate that the effect of reinforcement learning is unstable, but it can bring minor improvements, enhancing the model's coupling and accuracy. Downstream feedback was used to refine the upstream strategies. Tab. \ref{tab:compare_with_other_method} presents a quantitative comparison of the generation quality of several effective methods and their similarity to the real distribution. Additionally, comparison results with a broader range of models are available on the demo page and in the supplementary. The experimental findings indicate that our method excels in generating higher-quality videos. As shown in Fig. \ref{fig:demo1} and Fig. \ref{fig:demo2}, we visualize the effect of controlling the camera trajectory via text and controlling the camera trajectory via video.

\subsection{General Discussion and Human Study}
We conduct a comparison of state-of-the-art methods across various domains, examining the potential of different technical routes and highlighting the shortcomings of reconstruction methods, 4D reconstruction, and other approaches. Given the significant functional differences among models in various fields, adopting a unified quantitative evaluation metric is neither fair nor feasible. Therefore, we employe a manual evaluation method, inviting 50 participants to rate the methods on a scale of 1 to 5, with the final scores rounded. For Boolean evaluations, such as whether or not it is open source, open source is rated 5 out of 5 and not open source is rated 1 out of 5. The results are shown in Fig \ref{fig:radar}. The experimental findings indicate that ViewCrafter \cite{ViewCrafter} is cumbersome to interact with, especially when handling complex instructions, and it does not support learning camera trajectories from videos. ZeroNVS \cite{zeroNVS} is a novel view synthesis algorithm, but it can only generate one frame at a time and is cumbersome to use. GenWrap \cite{GenWarp}, another novel view generation algorithm, offers fast inference speed but suffers from generalization issues. CAT4D \cite{cat4d}, a 4D model, is hindered by its slow speed and lack of open-source availability. Additionally, reconstruction methods like One-2-3-45++ \cite{One-2-3-45++} are designed for single-object reconstruction and do not include scenes, so they are not considered in our study.

\section{Conclusion}
OmniCam is a unified multimodal camera control framework for video generation. It generates videos that meet user expectations by receiving text and video as trajectory references, and images and videos as content references. We utilize LLM to extract input features, obtain camera motion trajectories through trajectory planning algorithms, and finally obtain complete videos through 3D reconstruction and diffusion models. To support the full-process training of OmniCam, we have constructed the OmniTr dataset - the first multimodal dataset specifically designed for camera control. Experimental results show that our model demonstrates excellent robustness when faced with different modal combination inputs, and can accurately generate camera trajectory videos that conform to user intentions.

%% file: sec/X_suppl.tex
\clearpage
\setcounter{page}{1}
\maketitlesupplementary

\begin{figure*}[!htbp]
\centering
\includegraphics[width=\textwidth]{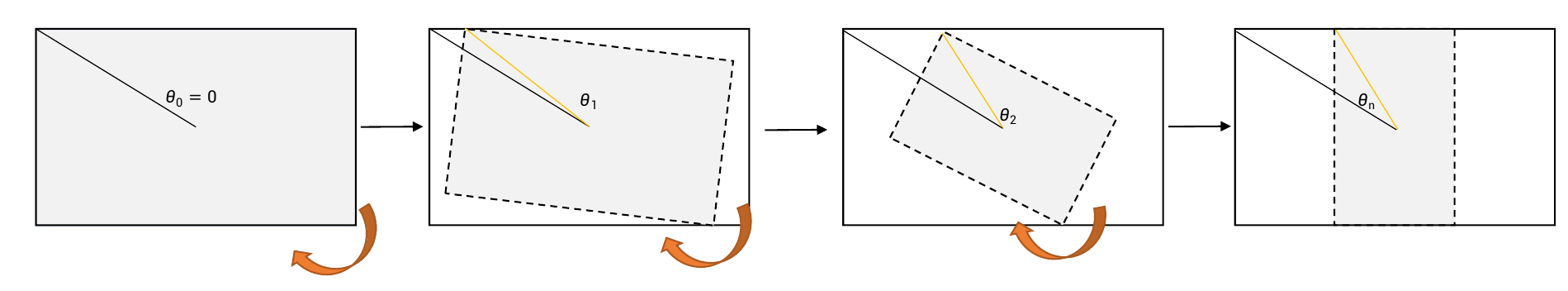} 
\caption{Visualization of Rotation Algorithm.}
\label{fig:rotate}
\end{figure*}

\section{Appendix}
\label{sec:rationale}
\subsection{Rotation Algorithm}
As shown in Fig. \ref{fig:rotate}, given a total rotation angle $\Theta$ (positive indicates counterclockwise rotation), the incremental rotation angle for each frame is $\Delta \theta=\frac{\Theta}{N}$, thus achieving a rotation angle of $\theta_k=k\times \Delta \theta$ at the $k$ frame. Assuming a rectangular image with dimensions $(w,h)$, after rotating around the center point $(\frac{w}{2},\frac{h}{2})$, its projected width and height can be expressed as:
\begin{equation}
    W(\theta)=w \cdot |cos\theta|+h \cdot |sin\theta|,
\end{equation}
\begin{equation}
    H(\theta)=w \cdot |sin\theta|+h \cdot |cos\theta|.
\end{equation}

To ensure the rotated image maintains the original aspect ratio $(w \times h)$ without distortion, a scaling factor $s$ must be applied to the image, satisfying:
\begin{equation}
    W(\theta)\times s \leq w,\quad H(\theta)\times s \leq h.
\end{equation}
Therefore, we can derive:
\begin{equation}
    s_w = \frac{w}{W(\theta)}=\frac{w}{w \cdot |cos\theta|+h \cdot |sin\theta|},
\end{equation}
\begin{equation}
    s_h = \frac{h}{H(\theta)} = \frac{h}{w \cdot |sin\theta|+h \cdot |cos\theta|},
\end{equation}
\begin{equation}
    s = min(s_w,s_h).
\end{equation}

The visual frequency is adjusted according to the input timing, enabling smooth visual rotation. After generating all keyframe data $(\phi,\theta,r)$, the algorithm unifies them into a single trajectory sequence.
\subsection{Metrics}
\noindent \textbf{CLIPSR} (CLIP-based Semantic Consistency Score) leverages the image encoding capabilities of the CLIP model to extract semantic features from each video frame and assesses the semantic coherence of the video by computing the similarity between frame features.

\noindent \textbf{NIQE} (Natural Image Quality Evaluator) is a no-reference image quality assessment method that evaluates image quality by extracting Natural Scene Statistics (NSS) features from the image and modeling them using a Multivariate Gaussian (MVG) model. It calculates the Mahalanobis distance between the test image features and the high-quality image statistical features. A lower score indicates a more natural image.

\subsection{More Related Work: Novel View Synthesis}
The process of transferring camera motion from video to static images is essentially a novel view synthesis problem. NeRF \cite{nerf} utilizes neural networks to learn volumetric scene information from multiple viewpoints, predicting color and density for each spatial point through network parameter optimization to achieve realistic view synthesis. 3D Gaussian Splatting \cite{3DGS} employs Structure-from-Motion (SfM) to obtain point clouds from multiple images, achieving near real-time 3D scene rendering by representing each point as a volumetric element (splat) with Gaussian distribution. While these methods typically require multiple images as input, users often can only provide a single image for creation. Consequently, researchers have begun exploring single-image novel view synthesis. Zero123 \cite{zero123} achieves new viewpoint image generation based on given camera poses by training diffusion models on synthetic datasets. TGS \cite{TGS} converts input images into Tri-plane feature representations and leverages 3D Gaussian Splatting for novel view rendering. However, these methods are limited to objects and consistently fail to generate 3D scenes. Recently, ZeroNVS \cite{zeroNVS} has achieved zero-shot novel view synthesis from a single input image through training on mixed datasets. Nevertheless, it struggles to synthesize consistent novel views and lacks precise pose control due to its treatment of camera pose conditions as high-level text embeddings. GenWarp \cite{GenWarp} combines text-image models with monocular depth estimation methods to generate new views, but its reconstruction remains unstable. ViewCrafter \cite{ViewCrafter} introduces a novel view synthesis framework that integrates video diffusion models with point cloud priors, achieving a balance between efficiency and fidelity.

\subsection{More Dataset Details}
The OmniTr dataset was initially created by using GPT to generate a large number of trajectory descriptions, which were then manually annotated with discrete motion representation information. Subsequently, these were converted into trajectory sequences in polar coordinate form through a trajectory planning algorithm, followed by the generation of corresponding videos using Viewcrafter. This data was later used for downstream training of OmniCam, essentially allowing OmniCam's downstream module to distill ViewCrafter's capabilities and further enhance them.

\begin{figure*}[!htbp]
\centering
\includegraphics[width=\textwidth]{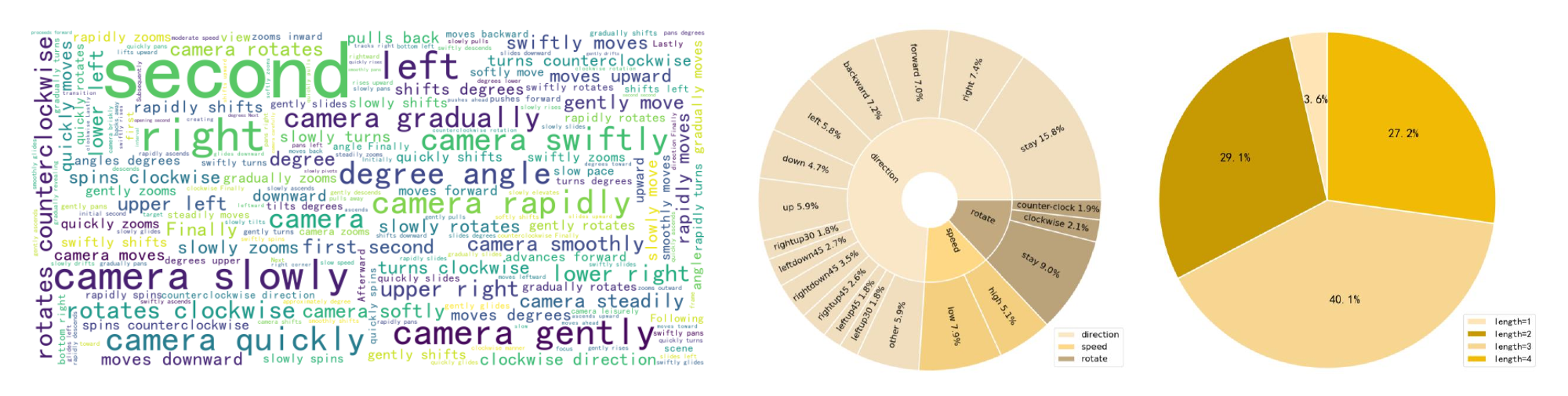} 
\caption{Visual analysis of the dataset.}
\label{fig:word_cloud}
\end{figure*}

As shown in the Fig. \ref{fig:word_cloud}, we conducted a visual analysis of the dataset to display data distribution, including word frequency distribution and motion feature statistics. The word cloud map shows that the keywords in the dataset mainly include time words (such as "second"), direction words (such as "right", "left", "upward", "downward"), speed words (such as "slow", "quick"), and motion words (such as "moves", "shifts", "rotors"). 
The first pie chart statistics show that single directional motion accounts for \textbf{27.5\%}, compound directional motion accounts for \textbf{25.7\%}, angular motion accounts for \textbf{40.1\%}, and rotational motion accounts for the remaining \textbf{6.7\%} in the dataset. The second pie chart shows the number of instructions for each group of texts. Most of the texts contain 2 to 4 groups of instructions.

This distribution reflects the comprehensive coverage of the dataset on different types of motion, especially the high proportion of angular motion and single-directional motion, which is consistent with practical application scenarios. Meanwhile, the balanced proportion between combined directional movements and rotational motions ensures the diversity and equilibrium of the dataset.

\subsection{More Training Details}
OmniCam employs a three-stage training strategy. In the first stage, we train a large language model using Llama 3.1 as the backbone. If text provides trajectory information, it is directly input into the LLM; if trajectory information comes from video, the video first undergoes SLAM to obtain preliminary pose estimation information. To ensure smoothness between different frames, we adopt a progressive inference approach, followed by using three MLPs to encode phi, r, and theta respectively, thereby mapping the pose sequence to our defined coordinate system.

In the second stage, we train the Video Diffusion Model. Following experimental settings similar to ViewCrafter but with improvements under the Dynamicrafter framework, we conducted 50,000 iterations of training with a learning rate of 5e-5 and a batch size of 16. Training was performed on subsets of the RealEstate, DL3DV, and OmniTr datasets, with each video containing 25 frames. We use DUSt3R to generate point cloud information, and during inference, we employ the DDIM sampler and classifier-free guidance.

In the third stage, we fine-tune the trajectory generation large model using reinforcement learning. Specifically, we extract trajectory information from downstream-generated videos using the video feature extractor trained in stage one, and convert the trajectory information in polar coordinate form into discrete-motion-representation through reverse parsing code. We then calculate Mavg as a reward, simulating human feedback, and fine-tune the large model using the RLHF (Reinforcement Learning from Human Feedback) method.

\subsection{More Result}
\begin{figure*}[!htbp]
\centering
\includegraphics[width=\textwidth]{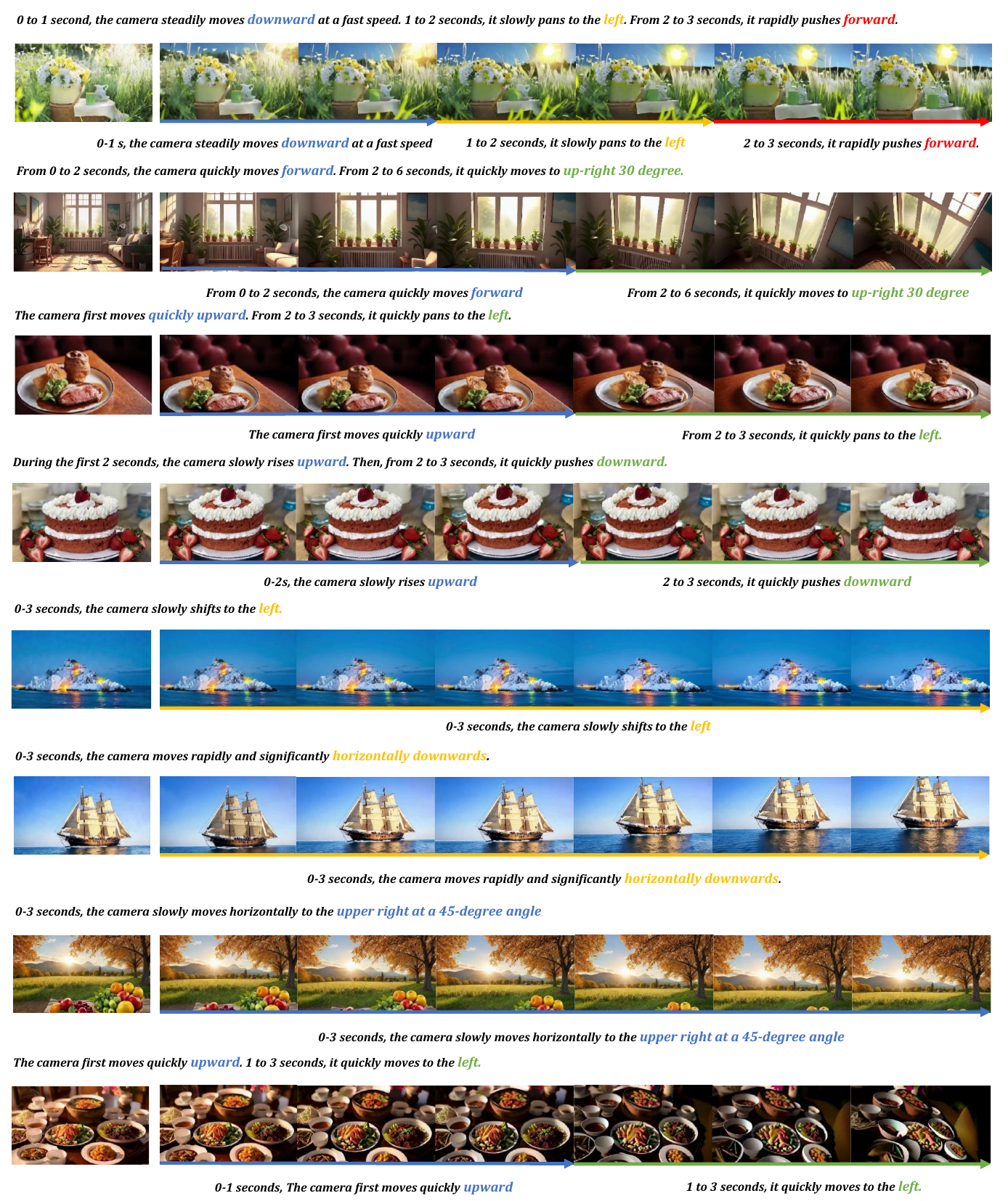} 
\caption{Some additional examples generated by OmniCam.}
\label{fig:merge_demo}
\end{figure*}

 We provide more generation results of our model in Fig. \ref{fig:merge_demo}. When confronted with different types of content references, our model can generate high-quality videos that follow the trajectory description.

\subsection{Comparison}
Recent years have witnessed a flourishing development in the field of video generation, with numerous remarkable research outcomes emerging. As this field is still in its exploratory stage without unified evaluation standards or mature paradigms, it is challenging to conduct comprehensive and fair comparisons of various models solely through quantitative metrics. Here, we present a brief comparative analysis based on our practical usage experience to help understand their characteristics and limitations.

\textbf{CAT4D} \cite{cat4d}: A closed-source 4D reconstruction algorithm developed based on CAT3D. Due to its closed-source nature, specific details and performance are difficult to fully understand.

\textbf{DimensionX} \cite{DimensionX}: As a 4D reconstruction model, DimensionX currently only supports image input with limited control capabilities, allowing only basic control in left-right directions without adjusting degree, time, or speed parameters. Additionally, its generation speed is less than half that of OmniCam, and it does not support secondary editing of videos.

\textbf{DynamiCrafter} \cite{DynamiCrafter}: A tool focused on transforming static images or text descriptions into high-quality dynamic videos, with satisfactory dynamic content generation effects.

\textbf{ViewCrafter} \cite{ViewCrafter}: This novel view generation model based on DynamiCrafter can generate perspective videos according to input trajectories; however, its capabilities in dynamic content generation are relatively limited.

\textbf{CogVideoX} \cite{CogVideoX}: An excellent open-source video tool supporting text-to-video generation and image control, but lacking support for video input and trajectory migration.

\textbf{CameraCtrl} \cite{CameraCtrl}: An excellent camera control tool trained on the RealEstate dataset, providing important inspiration for subsequent research. However, it lacks generalizability. Our experimental results show that camera trajectories selected from the RealEstate dataset perform well (the official inference code loads trajectories from this dataset). Since the RealEstate dataset contains many forward-moving camera effects, it performs well in forward and small-range backward movements. However, when we customize trajectories (such as "backward and right"), the spatial structure is compromised. Technically, it is less challenging for the camera to move to both sides while moving forward because the input image itself contains complete picture information, requiring less unknown information to be supplemented when moving forward and then right. Nevertheless, CameraCtrl is still excellent work, and we appreciate its contributions to the community.

\textbf{MotionMaster} \cite{MotionMaster}: This model can only provide content references through text and cannot control images. Its effects are demonstrated on the demo page.

\textbf{CAMI2V} \cite{CamI2V}: This model is designed for image-to-video conversion. By inputting a reference image and camera trajectory parameters (or text), it outputs videos that follow specified camera movements but currently does not support complex operations such as rotation and camera zooming.

\textbf{CineMaster} \cite{CineMaster}: Plans camera movement in 3D space and provides rich control capabilities, but the 3D operation workflow may still require some learning time for users unfamiliar with 3D modeling concepts. It also does not support video-guided trajectory generation.

\textbf{3DTrajMaster} \cite{3DTrajMaster}: Primarily focuses on controlling object movement rather than camera motion, which limits its application in video generation.

\textbf{RealCaM} \cite{RealCam}: Performs poorly when input images are not realistic enough, limiting its practical application effects.

\textbf{Lucid-Dreamer} \cite{liang2024luciddreamer}: Its results show serious artifacts because it uses depth-based deformation to generate new perspectives, which is particularly problematic when processing wild images (with unknown camera intrinsics), leading to inaccurate new perspective generation. Additionally, it adopts a ready-made repair model [43] to optimize deformation results, but this often introduces inconsistencies between original and repaired content.

\textbf{ZeroNVS} \cite{zeroNVS}: The quality of its generated new perspectives is relatively low with poor accuracy; the main reason is that ZeroNVS introduces camera pose conditions into the diffusion model through text embedding, which fails to provide precise control over new perspective generation, resulting in unsatisfactory results.

\textbf{MotionCtrl} \cite{MotionCtrl}: Can generate new perspectives with higher fidelity but performs inadequately in generating new perspectives precisely aligned with given camera conditions. This is because MotionCtrl also adopts high-level camera embeddings to control camera poses, resulting in lower accuracy in new perspective synthesis.

\textbf{DNGaussian} \cite{DNGaussian}: A neural rendering method based on Gaussian distribution, aiming to generate high-quality scenes through probabilistic models. However, its results exhibit significant artifacts, indicating limited rendering capabilities in complex scenes.

\textbf{FSGS} \cite{FSGS}: A fast scene generation method based on sparse Gaussian distribution, focusing on efficiently generating novel views. However, its results also show artifacts when viewed from novel perspectives that deviate from the ground truth training images, suggesting insufficient robustness in view extrapolation tasks.

\textbf{InstantSplat} \cite{InstantSplat}: A real-time scene generation method based on point clouds, utilizing DUSt3R for point cloud initialization, which better preserves details from the ground truth training images. However, due to its omission of the densification process, it fails to recover occlusion regions, resulting in incomplete performance in complex scenes.

Simultaneously, most works (such as CameraCtrl, RealCaM, and CAMI2V) are trained on the RealEstate dataset, lacking full-process training for trajectory control. This limitation affects their generalization ability and application in broader scenarios.

Furthermore, we also compared with some commercially available large models that have been launched:

\textbf{Hunyuan} \cite{Hunyuan}: It has the powerful ability to generate videos from text, but it does not support providing trajectories through videos. Users need to input the corresponding camera trajectories in text form, and only simple zoom-in and zoom-out operations can be achieved.

\textbf{Tongyi} \cite{tongyiwanxiang2024}: Has strong text-to-video generation capabilities but lacks spatial awareness, limiting its flexibility.

\textbf{Runway} \cite{RunwayGen2}: An excellent commercial perspective control model with the best spatial consistency in benchmark tests. It supports image input and controls simple operations (such as "room in," "right," etc.) through buttons, allowing for the superposition of multiple directions but not supporting continuous multiple operations and time control.

Specifically, we compared the input modalities of multiple models. As shown in the Tab. \ref{tab:comparison}, our model covers all input scenarios, facilitating various types of creation for users.

\begin{table}
\centering
\caption{Comparison of Content and Trajectory Reference Capabilities}
\resizebox{\columnwidth}{!}{
\begin{tabular}{lcccccc}
\toprule
\multicolumn{1}{c}{\textbf{Model}} & \multicolumn{3}{c}{\textbf{Content Reference}} & \multicolumn{3}{c}{\textbf{Trajectory Reference}} \\
\cmidrule(lr){2-4} \cmidrule(lr){5-7}
 & \textbf{Text} & \textbf{Image} & \textbf{Video} & \textbf{Text} & \textbf{Video} & \textbf{Trajectory} \\
\midrule
\centering MotionMaster & \ding{51} & & & & \ding{51} & \\
\centering OmniCam & \ding{51} & \ding{51} & \ding{51} & \ding{51} & \ding{51} & \ding{51} \\
\centering CameraCtrl & \ding{51} & \ding{51} & & & \ding{51} & \ding{51} \\
\centering CogVideoX & \ding{51} & & & \ding{51} & & \\
\centering DimensionX & & \ding{51} & \ding{51} & & & \ding{51} \\
\centering 3DTrajMaster & \ding{51} & & & & & \ding{51} \\
\centering RealCam & \ding{51} & \ding{51} & & \ding{51} & & \ding{51} \\
\centering CineMaster & \ding{51} & & & & & \ding{51} \\
\centering CAMI2V & \ding{51} & \ding{51} & & \ding{51} & & \ding{51} \\
\centering Runway & & \ding{51} & & \ding{51} & & \\
\centering Hunyuan & \ding{51} & & & \ding{51} & & \\
\centering Tongyi & \ding{51} & & & \ding{51} & & \\
\bottomrule
\end{tabular}}
\label{tab:comparison}
\end{table}

\subsection{Details Concerning Table 3}

\begin{figure}[t]
\includegraphics[width=\columnwidth]{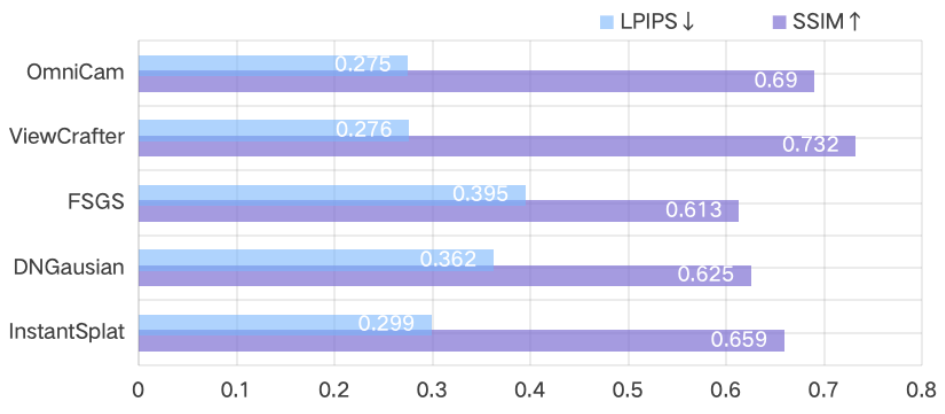} 
\caption{Comparison of performance of OmniCam, ViewCrafter, FSGS, DNGaussian, InstantSplat and other methods on two indicators: LPIPS and SSIM.}
\label{fig:comparison}
\end{figure}

Since some methods \cite{zeroNVS,liang2024luciddreamer} are only capable of processing square images, we crop the generated novel views from our method and MotionCtrl to ensure alignment when computing the quantitative metrics. On our demo page, we present extensive comparisons with related works. Additionally, to further validate the novel view synthesis capabilities of our model from a 3D reconstruction perspective, we conducted experiments on a subset of the Tanks-and-Temples \cite{Knapitsch2017} dataset. Specifically, we used 2 ground truth training images for each scene and evaluated the performance using 12 novel views. Its results are shown in Fig. \ref{fig:comparison}. Currently, most reconstruction models focus primarily on object reconstruction, while research on scene reconstruction remains relatively limited. To comprehensively evaluate scene reconstruction capabilities, we selected several representative scene reconstruction models for comparative experiments. The experimental results demonstrate that DNGaussian exhibits significant artifacts in complex scenes, particularly in high dynamic range (HDR) or geometrically complex structures, where issues such as detail loss and edge blurring frequently occur. FSGS performs poorly in view extrapolation tasks; when the viewpoint deviates from the ground truth training images, the generated artifacts increase significantly, making it difficult to maintain geometric consistency and texture continuity. InstantSplat, which utilizes DUSt3R for point cloud initialization, better preserves details from the ground truth training images. However, due to its omission of the densification process, it fails to effectively recover occluded regions, resulting in holes or distortions in complex scenes and compromising the overall visual quality. These limitations highlight the challenges in novel view synthesis tasks, particularly in achieving geometric consistency, texture continuity, and effective occlusion handling.

